# Universal crystal material property prediction via multi-view geometric fusion in graph transformers

Liang Zhang[1, 2], Kong Chen[1, 2, *] and Yuen Wu[1, 2, *]

[1]Key Laboratory of Precision and Intelligent Chemistry/School of Chemistry and Materials Science, University of Science and Technology of China, Hefei, China.
[2]National Key Laboratory of Deep Space Exploration, Deep Space Exploration Laboratory, Hefei, China.

[*]Corresponding authors. E-mail: yuenwu@ustc.edu.cn; kongchen@ustc.edu.cn

# Abstract

Accurately and comprehensively representing crystal structures is critical for advancing machine learning in large-scale crystal materials simulations, however, effectively capturing and leveraging the intricate geometric and topological characteristics of crystal structures remains a core, long-standing challenge for most existing methods in crystal property prediction. Here, we propose MGT, a multi-view graph transformer framework that synergistically fuses SE3 invariant and SO3 equivariant graph representations, which respectively captures rotation-translation invariance and rotation equivariance in crystal geometries. To strategically incorporate these complementary geometric representations, we employ a lightweight mixture of experts router in MGT to adaptively adjust the weight assigned to SE3 and SO3 embeddings based on the specific target task. Compared with previous state-of-the-art models, MGT reduces the mean absolute error by up to 21% on crystal property prediction tasks through multi-task self-supervised pretraining. Ablation experiments and interpretable investigations confirm the effectiveness of each technique implemented in our framework. Additionally, in transfer learning scenarios including crystal catalyst adsorption energy and hybrid perovskite bandgap prediction, MGT achieves performance improvements of up to 58% over existing baselines, demonstrating domain-agnostic scalability across diverse application domains. As evidenced by the above series of studies, we believe that MGT can serve as useful model for crystal material property prediction, providing a valuable tool for the discovery of novel materials.

# Introduction

Crystal materials are widely used in areas such as energy storage, catalysis, drug discovery and semiconductor design due to their unique structures and properties[1-4]. Accurate prediction of the properties of crystal materials is crucial for accelerating the discovery of new materials with specific functions[5, 6]. However, traditional methods such as Density Functional Theory (DFT) calculations or molecular dynamics (MD) simulations are both time-consuming and costly. The computational complexity of DFT scales cubically with the system size, which requires significant computational resources and time, especially when dealing with material systems containing a large number of atoms[7-9].

In recent years, significant progress has been made through the development of graph neural network (GNN)-based models, enabling rapid and accurate crystal material property prediction. In 2018, Xie et al. introduced the CGCNN framework, capable of learning material properties directly from atomic connections in crystal graphs[10]. Since then, a variety of GNN-based models have emerged. For instance, Matformer[11], a graph transformer designed to be invariant to periodicity, explicitly captures repeating patterns and characterizes crystal structures through geometric distances between identical atoms in neighboring cells. PotNet[12] innovatively models the complete set of interatomic potentials among all atoms by computing physics-principled potentials, such as the Coulomb, London dispersion, and Pauli repulsion potentials, achieving the infinite summations of pairwise atomic interactions. However, employing graph representations that only consider local information for crystal structures cannot guarantee a one-to-one correspondence between the representation and the crystal graph[13]. This limitation makes it difficult for GNN models to distinguish distinct crystal materials, thus impacting prediction accuracy.

To address this challenge, PerCNet[13] incorporates dihedral angles to resolve the many-to-one problem in crystal graph representations. Ablation experiments confirm that this additional information is essential for the geometric completeness of periodic structures. Specially, in crystal materials, atoms are arranged in a periodic lattice in three-dimensional (3D) space, where the geometric structure is SE3 invariant or SO3 equivariant under specific translations and rotations[14]. These properties reflect the underlying spatial symmetries of the crystal, corresponding to symmetry operations of a subgroup of the Euclidean group E3. Indeed, recent works based on SE3 invariant and SO3 equivariant have already shown great potential in atomic systems[15-17]. Equiformer[16] leverages the Transformer architecture[58] to achieve strong empirical results on 3D atomic benchmarks such as QM9, MD17 and OC20, highlighting the advantage of Transformer-based model with equivariant features in atomistic systems. ComFormer[14] introduces two variants, iComFormer and eComFormer, by incorporating SE3 invariant and SO3 equivariant crystal graph representations, respectively. As a result, ComFormer demonstrates excellent performance across multiple crystal property prediction tasks, underscoring the efficiency and expressiveness of SE3 invariant and SO3 equivariant graph features.

Although there have been significant advancements in leveraging SE3 invariant and SO3 equivariant graph features, current state-of-the-art models still fall short of fully harnessing their potential. For instance, prior works have predominantly focused on utilizing these features in isolation within neural frameworks, neglecting their intrinsic synergistic capabilities. Furthermore, while one[18] has incorporated domain knowledge and rules of dual-graph features, it primarily relies on a simple concatenation method that fail to exploit the advantages of the leveraged knowledge. Consequently, the absence of an integrated framework to effectively fuse multi-view features and maximize their synergistic benefits remains a significant unresolved challenge[19], highlighting the need for a novel approach that can unlock their full potential and significantly enhance the accuracy and efficiency of crystal material property prediction.

In this work, we propose a novel framework, dubbed MGT (Multi-view Graph Transformer), designed to fully exploit the comprehensive and accurate geometric features of both SE3 invariance and SO3 equivariance. We leverage pretraining and fine-tuning strategies[20, 21] with MGT. Specifically, a multi-task self-supervised learning[22, 23] (SSL) pretraining strategy, including denoising and contrastive learning, is employed to enhance the model's capability to effectively recognize crystal structures. In the fine-tuning stage, we introduce a lightweight mixture of experts (MoE)[24, 25] module within MGT to dynamically and efficiently process SE3 and SO3 embeddings, which meaningfully improves the performance of the model in downstream tasks. We compare MGT with a series of models for crystal material property prediction and conduct ablation studies to systematically evaluate the contribution of each component within MGT. Our results confirm that MGT outperforms baseline models across multiple prediction tasks, with performance improvements of 3.3-20.8%. Our illustrations demonstrate that MGT has effective clustering in the latent space, as revealed by t-Distributed Stochastic Neighbor Embedding (t-SNE)[28] visualization. Furthermore, we apply MGT to predict the global minimum adsorption energy (GMAE) of catalysts[26] and the bandgap of hybrid organic inorganic perovskites (HOIP)[27] through transfer learning. Compared to state-of-the-art models in these domains, MGT achieves performance gains ranging from 15.4% to 57.8%. Our main contributions in this work are as follows.

- We develop a multi-view graph transformer framework named MGT for crystal property prediction, consisting of two graph transformer encoders and a light MoE router to efficiently process SE3 invariant and SO3 equivariant graph features.
- We introduce an SSL pretraining strategy for MGT that incorporates denoising and contrastive learning in the pre-training stage. The effectiveness of this pretraining approach is validated through ablation experiments.
- We demonstrate that MGT can effectively distinguish distinct crystal materials in the latent space. Further, these insights elucidate the underlying factors for the superior performance of MGT.
- We successfully implement MGT in catalyst GMAE and HOIP bandgap prediction tasks, achieving competitive prediction performance and demonstrating its generalizability and scalability in practical applications.

## Results and discussion

**Preliminaries**

**Crystal representation.** A crystal material $M$ can generally be represented as a periodic arrangement of atoms within a unit cell[29]. Formally, give $M = (A, X, L)$, where the infinite periodic structure can be expressed as:

$$\{(\hat{a}_i, \hat{x}_i) | \hat{a}_i = a_i, \hat{x}_i = x_i + k_1 \ell_1 + k_2 \ell_2 + k_3 \ell_3, \ k_1, k_2, k_3 \in \mathbb{Z}, 1 \leq i \leq n\} \quad (1)$$

where $A = [a_1, a_2, ..., a_n] \in \mathbb{A}^N$ denotes the list of atomic species, $X = [x_1, x_2, ..., x_n] \in \mathbb{R}^{N \times 3}$ is the corresponding atomic coordinates, $L = [\ell_1, \ell_2, \ell_3] \in \mathbb{R}^{3 \times 3}$ is the lattice matrix defining the periodic translations in three dimensions and $k_1, k_2, k_3$ are integer translation indices along the respective lattice directions, used to extend the unit cell periodically in 3D space.

**SE3 and SO3.** A crystal representation $f: (A, X, L) \to y$ preserve fundamental geometric symmetries such as SE3 invariance and SO3 equivariance, to ensure rotational consistency and geometric completeness. The representation is SE3 invariant if it satisfies: $f(A, X, L) = f(A, XR^T + b, RL)$, for any rotation matrix $R \in \mathbb{R}^{3\times 3}$, $|R|=1$, and translation vector $b \in \mathbb{R}^3$. This ensures that the output remains unchanged under rotations and translations applied simultaneously to atomic coordinates $X$ and lattice vectors $L$. The representation is SO3 equivariant if it satisfies: $f(A, XR^T, RL) = f(A, X, L)R$, for any rotation matrix $R \in \mathbb{R}^{3\times 3}$, $|R|=1$. This ensures that vectorial outputs rotate consistently with the input.

**Graph construction.** To capture symmetries and periodicity, crystal graphs are constructed by associating each atom $i$ with periodic images $j'$ of other atoms $j$. The position of each periodic image $x_{j'}$ is defined as:

$$x_{j'} = x_i + k_1 \ell_1 + k_2 \ell_2 + k_3 \ell_3, \ k_1, k_2, k_3 \in \mathbb{Z} \tag{2}$$

An edge is formed between atom $i$ and a periodic image $j'$ if the Euclidean distance satisfies: $\left\| x_{j'} - x_i \right\|_2 \leq r$, where $r$ is a predefined cutoff radius. For SE3 invariance graphs, edges incorporate scalar geometric features, including Euclidean distances $\left\| x_{j'} - x_i \right\|_2$ and angles $\theta_{j'i,ii_k}, k \in \{1, 2, 3\}$ between edge vectors $x_{j'} - x_i$ and lattice reference vectors $\{e_{ii_1}, e_{ii_2}, e_{ii_3}\}$, which are constructed from the closest periodic images of atom $i$ and preserve the periodic invariance. For SO3 equivariant graphs, edges include both the scalar distances $\left\| x_{j'} - x_i \right\|_2$ and the equivariant vectors $x_{j'} - x_i$, which transforms under rotation. Moreover, the geometric completeness of SE3 invariant and SO3 equivariant crystal graphs respects both lattice periodicity and geometric symmetries, which can uniquely determine a crystal structure, as demonstrated in prior work[14].

**Framework**

As shown in Fig. 1, MGT is composed of two graph encoders and a MoE module, designed to process SE3 invariant and SO3 equivariant representations, respectively. Each graph encoder incorporates a graph transformer architecture and a lightweight projection head. Specifically, the SE3 graph encoder employs both edge-wise and node-wise transformer layers to update node, edge, and angle features via attention-based message passing, effectively capturing SE3 invariant geometric information. In contrast, the SO3 graph encoder leverages spherical harmonics and tensor product (TP) layers to ensure

SO3 rotational equivariance while enriching structural representations. Additionally, to leverage both SE3 invariant and SO3 equivariant features during the fine-tuning on downstream tasks, The lightweight MoE module adaptively integrates SE3 and SO3 embeddings via a self-attention-based router[25], enabling task-specific fusion of geometry-aware representations.

Recent studies in representation learning have demonstrated that SSL pretraining can significantly enhance the generalization of models on downstream tasks[30-32]. Admittedly, to enhance the effectiveness of model and reduce its reliance on large labelled datasets, MGT also adopts a multi-task SSL pretraining strategy tailored to the unique challenges of crystal property prediction. This strategy includes two distinct pretraining strategies: denoising learning and contrastive learning.

For the denoising learning strategy[33], Gaussian noise is added to angle and edge features for SE3 and SO3 representations, respectively. Following the approach of GPIP framework[66], MGT is trained to restore the disturbed noise, rather than reconstructing the original features. This training strategy forces MGT to learn invariant and equivariant features that remain stable under local perturbations, while preserving geometric symmetries. For the contrastive learning strategy[34], MGT aligns SE3 invariant with SO3 equivariant representations, maximizing the mutual information across these geometric features. This contrastive strategy forces the model to learn complementary features from each representation, enhancing the model's capability to generalize across various crystal structures. The combination of these two pretraining strategies enables MGT to capture meaningful spatial structural features and enhance cross-view consistency between SE3 and SO3 embeddings.

**Experiments**

**Performance evaluation.** We evaluated the performance of MGT framework on nine downstream tasks using two benchmark crystal material datasets: Materials Project[35] and JARVIS[36]. The evaluation was compared against several advanced baselines, including CGCNN[10], ALIGNN[37], Matformer[11], PotNet[12], ComFormer[14] and ReGNet[38]. To ensure a fair comparison of model performance, we also employed the same data splitting seed as in prior studies, with the mean absolute error (MAE) and $R^2$ serving as the assessment criterion. As shown in Supplementary Tables 1 and 2, MGT outperforms these baseline methods in eight downstream prediction tasks, reducing MAE ranging from 3.3% to 20.8%. Notably, as shown in Fig. 3, MGT demonstrates robust fitting performance across all downstream tasks, with the predicted data points predominantly clustering around the diagonal and all $R^2$ values exceeding 0.95. These results demonstrate the effectiveness of multi-view features combining SE3 invariance and SO3 equivariance in MGT, leading to improved performance in crystal material property prediction tasks.

**Ablation studies.** To study the impact of each technique we adopted on model performance, we carried out ablation experiments in three aspects, which are the performance comparison with and without pretraining, the effectiveness of each pretraining strategy and the performance comparison with and without the MoE module in downstream task fine-tuning. Firstly, we compared MGT with and without pretraining. As shown in Fig. 4a and Supplementary Table 3, MGT with pretraining achieves an MAE of 0.0165, which is better than the MAE of 0.0174 obtained without pretraining on the Formation Energy prediction task for Materials Project dataset. This demonstrates that the pretraining strategy is more effective than training MGT framework from scratch in achieving high performance.

Secondly, we studied MGT with the denoising learning strategy, the contrastive learning strategy and both strategies combined. As shown in Fig. 4 and Supplementary Table 3, MGT with both the denoising and contrastive learning strategies achieves the best model performance. However, it is worth noting that MGT with the denoising learning strategy achieves MAE of 0.116, outperforming MAE of 0.120 achieved by MGT with the contrastive learning strategy on the Bandgap (OPT) prediction task for JARVIS dataset. This performance trend also holds across both datasets in the Formation Energy prediction task. These findings suggest that the denoising pretraining strategy may be more critical than the contrastive learning strategy in crystal property prediction. By focusing on restoring noise, the denoising strategy enhances the perceptive ability of MGT to recognize SE3 invariant and SO3 equivariant geometric features. Furthermore, when the denoising pretraining is combined with the contrastive learning strategy, it improves the generalization of MGT across downstream prediction tasks, achieving the best performance.

Lastly, we also evaluated the performance of MGT with and without the MoE module. In the absence of the MoE module, we replaced it with a simple NN architecture[39]. Specifically, the features from both SE3 and SO3 embeddings were concatenated and then passed through two fully connected layers. A more detailed description is provided in Supplementary Note 1. As shown in Fig. 4 and Supplementary Table 3, MGT with the MoE module achieves MAEs of 0.0261 and 0.114, outperforming MAEs of 0.0263 and 0.115 achieved by MGT without the MoE module on the Formation Energy and Bandgap (OPT) prediction tasks for JARVIS dataset, respectively. MGT with the MoE module also outperforms the one without the MoE module on Materials Project dataset. Therefore, these results strongly demonstrate that the MoE module in MGT enhances model performance by effectively leveraging SE3 and SO3 embeddings and dynamically integrating these features during the fine-tuning phase. This adaptive capability offers a clear advantage over traditional NN architectures in fusing dual representations. Moreover, we can also see that MGT still achieves better performance than baseline methods even without pretraining or MoE module, demonstrating the inherent effectiveness of multi-view architecture alone already outperforms the baselines.

**Interpretability analysis.** To effectively assess the contributions of each component within MGT framework, we utilized t-SNE[28] for visualization analysis with SE3, SO3 and MoE embeddings. Details can be found in Supplementary Note 2. As illustrated in Fig. 5a-d, the SE3 and SO3 graph encoders successfully cluster crystal materials based on the Formation Energy for Materials Project and JARVIS datasets. Furthermore, as illustrated in Fig. 5e and 5f, the MoE module strengthens the clustering, making the groupings more distinct. This improved clustering is crucial for MGT to distinguish distinct crystal materials. The clustering effectiveness is consistent across all prediction tasks, as shown in Supplementary Fig. 1-7. These results highlight the significant role of the MoE module in enhancing overall model performance on downstream tasks.

Next, we also analyzed the actual contribution scores of SE3 and SO3 embeddings in the MoE module. Details are provided in Supplementary Note 3. As shown in Extended Data Fig. 1a, in the Formation Energy prediction task, SO3 embeddings exhibit a more significant contribution than SE3 embeddings for Materials Project dataset. However, as shown in Extended Data Fig. 1b, the opposite is observed for JARVIS dataset, where SE3 embeddings contribute more. Additionally, as shown in Extended Data Fig. 1c, SE3 and SO3 embeddings contribute almost equally in the Band Gap prediction task for Materials Project dataset. In contrast, as shown in Extended Data Fig. 1d, the Bandgap (OPT) prediction task for JARVIS dataset exhibits negative and positive contribution scores for SE3 and SO3 embeddings, respectively. These findings indicates that the MoE module within MGT can autonomously adjust the weight assigned to SE3 and SO3 embeddings depending on the specific prediction task. We also presented the contribution scores of other prediction tasks, as shown in Supplementary Fig. 12 and 13. The SE3 and SO3 graph encoders each demonstrate unique strengths in different tasks, and their weighting relationship appears to be complementary. For example, in the Bulk Moduli, Bandgap (OPT) and Bandgap (MBJ) prediction tasks for Materials Project and JARVIS datasets, respectively, this complementary nature is particularly evident. This dynamic adjustment mechanism of MoE module enables MGT to optimize the contributions of SE3 and SO3 embeddings according to different prediction tasks, thereby significantly improving model performance.

**Transfer learning.** It is essential to evaluate the effectiveness of MGT framework by assessing its generalizability and scalability through transfer learning[40, 41]. To this end, we applied MGT to two diverse transfer learning scenarios. The first involved predicting GMAE, a crucial reactivity descriptor that helps identify the most stable adsorption configurations of catalysts. Accurate prediction of GMAE is essential for the effectiveness of machine learning algorithms in catalyst screening. As shown in Fig. 6a-c, MGT demonstrates satisfactory fitting performance across the Alloy-GMAE, FG-GMAE, and OCD-GMAE datasets. Compared with the state-of-the-art AdsMT[26] baseline, as detailed in Supplementary Table 4, MGT achieves significant improvements of 46.4%, 57.8% and 15.4%, respectively. The visualizations and contribution scores for these prediction tasks also confirm our observations, as shown in Supplementary Fig. 8-10 and 14.

The second scenario focused on predicting the bandgap of HOIP[27], for which precise prediction is critical for optimizing optoelectronic applications such as solar cells and light-emitting diodes. As shown in Fig. 6d and Supplementary Table 5, MGT exhibits competitive performance compared to other baseline methods in the bandgap prediction task on the HOIP dataset, with a 25.7% improvement. We also presented all visualizations and contribution scores, as shown in Supplementary Fig. 11 and 14, respectively.

Overall, these findings demonstrate the excellent generalization ability and scalability of MGT framework across diverse transfer learning scenarios, highlighting its promising potential for practical applications.

## Conclusion

In conclusion, we have introduced MGT, a geometry-aware multi-view graph transformer that unifies SE3 invariant scalar geometry and SO3 equivariant vector geometry for universal crystal property prediction. Our extensive experiments have clearly demonstrated that MGT outperforms other algorithms in multiple prediction tasks, achieving highly accurate and reliable performance. This underscores the rationality and effectiveness of our multi-task pretraining strategy and the robustness of our architecture design. We have also applied MGT to transfer learning scenarios, where it has demonstrated superior performance compared to state-of-the-art models in GMAE and HOIP bandgap prediction tasks. Overall, our findings provide compelling evidence that MGT is an efficient and powerful framework for crystal material property prediction.

## Limitations and Outlook.

MGT currently focuses on SE3/SO3 symmetries of static structures and does not yet incorporate the full crystallographic space group, magnetic or time-reversal symmetries, nor finite-temperature or defect-induced disorder. Extending the framework to space-group-equivariant message passing, stochastic symmetry transformations and multi-fidelity active-learning loops will be key next steps[42-45]. Coupling MGT with symmetry-preserving diffusion or autoregressive models will further enable direct inverse design of crystals with target properties[46-48], paving the way toward autonomous, AI-accelerated discovery of next-generation functional materials.

## Methods

### Dataset

The pretraining dataset is sourced from the Open Quantum Materials Database (OQMD)[49], specifically version 1.5, which is released in October 2021. The data within this version had been optimized through density functional theory (DFT) simulations. In cases where multiple structures shared the same cell composition and space group, we carefully selected the structure that exhibited the smallest unit volume per chemical formula following Antunes et al[50]. This selection process ultimately yielded a dataset comprising 587483 structures, which are then partitioned into training, validation, and testing sets in a ratio of 8:1:1. Element distribution of OQMD was shown in Extended Data Figs. 4.

The downstream datasets used in our experiments are Materials Project[35] and JARVIS[36]. The Materials Project, version is 2018.6.1, which contains 69239 structures for the Formation Energy and Band Gap prediction tasks. However, for the Bulk Moduli and Shear Moduli prediction tasks, only 4664 structures are available. JARVIS includes 55722 structures for the Formation Energy, Bandgap (OPT), Total Energy prediction tasks. For the Bandgap (MBJ) and Ehull prediction tasks, JARVIS provides 18171 and 55370 structures, respectively. We evaluated model performance on Materials Project and JARVIS datasets, which involve four and five crystal property prediction tasks, respectively. All downstream datasets used the same training, validation and testing splits and data sizes, following the methodology of Yan et al.[11] and Lin et al.[12]. Element distributions of Materials Project and JARIVS were shown in Extended Data Figs. 2 and 3.

The datasets employed for transfer learning include Alloy-GMAE, FG-GMAE, and OCD-GMAE, which are derived from the Catalysis Hub[51], the 'functional groups' (FG) dataset[52], and the OC20-Dense dataset[53], respectively. Alloy-GMAE dataset comprises the largest number of structures, totaling 11260. FG-GMAE dataset features a medium scale with 3038 structures. OCD-GMAE dataset consists of 973 structures. HOIP dataset, made publicly available by Kim et al.[27], includes 1346 HOIP structural files. To ensure fair comparisons across models, we also utilized OC20-LMAE[54] as the pretraining dataset for Alloy-GMAE, FG-GMAE and OCD-GMAE as Chen et al.[26]. OC20-LMAE dataset was filtered from the OC20[54], which eventually comprises 363937 structures. The GMAE-related datasets, including the pretraining dataset, underwent rigorous data cleaning and organization as demonstrated by Chen et al.[26]. Each dataset was also split into training, validation and testing sets with a ratio of 8:1:1. A more detailed display is shown in Supplementary Table 6.

**MGT**

MGT is designed to capture the geometry information contained in the two distinct representations of SE3 and SO3 graphs. It is a dual-graph encoder architecture that processes node features, edge features, and angle features in the graphs, respectively. Specifically, the initial node features are derived from CGCNN embedding features[10]. For edge features, such as node $j$ to node $i$, the initial features are obtained using Radial Basis Function (RBF) kernel[55] embedding features. The initial angle features are based on RBF kernel embedding with the cosine function employed. The first encoder of MGT consists of the SE3 graph transformer and the SE3 projection head layer. As illustrated in Fig. 2a and 2b, the SE3 graph transformer includes edge-wise transformer and node-wise transformer layers[14] for updating correspondent messages. The edge-wise transformer layer updates edge features with angle information and crystal lattice basis vectors, which can be expressed as:

$$Q_{ij}^e = f_Q^e(e_{ij}),$$

$$K_{ij}^e = \phi_K^e \left( f_K^e(e_{ij}) \parallel f_{K_m}^l(\ell_m) \parallel f_E^a(\theta_{ij}) \right), \quad (3)$$

$$V_{ij}^e = \phi_V^e \left( f_V^e(e_{ij}) \parallel f_{V_m}^l(\ell_m) \parallel f_E^a(\theta_{ij}) \right)$$

where $f_Q$, $f_K$, $f_V$ and $f_E$ are linear transformation layers, $\phi_K$ and $\phi_V$ are nonlinear layers, $e_{ij}$ is the edge features, $\theta_{ij}$ is the angle features, and $\ell_m$ (m=3) are the lattice basis vectors of the unit cell. Then the attention weight computation and message aggregation output can be formulated as:

$$Output = Softplus(e_{ij} + BNorm\left( \sum_{i=1}^{m} \sigma\, BNorm\left( \frac{Q_{ij} K_{ij}}{\sqrt{d_{Q_{ij}}^k}} \right) V_{ij} \right)) \quad (4)$$

where $\sigma$ denotes the sigmoid function, $d_{Q_{ij}}^k$ is the output dimension of each query vector $Q_{ij}$, $Softplus$ denotes the activation function and $BNorm$ denotes the batch normalization. Then, the node-wise transformer layer aggregates messages from neighboring nodes and edge features, where the edge features are computed from the edge-wise transformer layer, which can be expressed as:

$$Q_i^h = f_Q^h(h_i),$$

$$K_{ij}^h = \phi_K(f_K^i(h_i) \parallel f_K^j(h_j) \parallel f_E^e(e_{ij})), \quad (5)$$

$$V_{ij}^h = \phi_V(f_V^i(h_i) \parallel f_V^j(h_j) \parallel f_E^e(e_{ij}))$$

$$Output = Softplus(h_i + BNorm\left( \sum_{i=1}^{n} \sigma\, BNorm\left( \frac{Q_i K_{ij}}{\sqrt{d_{Q_i}^k}} \right) V_{ij} \right)) \quad (6)$$

where $h_i$ is the node features, $d_{Q_i}^k$ is the output dimension of each query vector $Q_i$, and $n$ denotes the number of neighboring nodes connected to node $i$. The final component is a lightweight SE3 projection head layer, which can be formulated as:

$$z = z + LNorm(f_{l_2}((f_{l_1}z + b_1) + b_2)) \tag{7}$$

where $f_l$ is the linear layer and $LNorm$ is the layer normalization.

To effectively capture the SO3 equivariant features of crystal graphs, we also adopted an equivariant node update module within the SO3 graph encoder. As illustrated in Fig. 2c, This module is based on spherical harmonics[56] and TP operations following by Geiger et al.[57] and Yan et al.[14]. Specifically, node features are first projected to a scalar representation through linear transformations. Then, two TP layers are applied, each combining node features, edge-wise spherical harmonics and edge features. The TP layer updates the node features by computing a weighted TP of the neighboring node features $h_j$, the spherical harmonics $Y_{ij}$ derived from relative position vectors $r_{ij}$ and a learned edge-dependent weight $W(e_{ij})$. Formally, each TP layer is defined as:

$$h_i^{(l+1)} = Agg_{j \in N(i)} \left[ TP\left(h_j^{(l)}, Y_{ij}, W(e_{ij})\right) \right] + h_i^{(l)} \tag{8}$$

where $Agg$ is an aggregation function, and residual connections were also used to preserve representational depth. Then, the output feature is followed by a $Softplus$ activation, batch normalization and a linear projection layer, with another residual connection from the initial input feature $h_i^{(0)}$:

$$Output = Softplus\left(f_l \, Softplus\left(BNorm\left(h_i^{(2)}\right)\right)\right) + h_i^{(0)} \tag{9}$$

this architecture guarantees that the learned node representations remain equivariant to rotations, effectively capturing both scalar and higher-order geometric interactions through the TP operation. Similar to the SE3 encoder, we also incorporated a node-wise transformer followed by a lightweight SO3 projection head layer for the final output.

Furthermore, to effectively integrate SE3 and SO3 graph embeddings from different graph encoders for downstream tasks, as illustrated in Fig. 2d, we introduce a lightweight MoE[24, 25] module consisting of two parallel expert networks and a self-attention-based router that dynamically assigns importance weights to each expert. Given input embeddings $E_1$ and $E_2$ from the SE3 and SO3 graph encoders, respectively, the fused representation is computed as:

$$H = [f_{expert1}(E_1), f_{expert2}(E_2)],$$
$$W_{router} = Router_{attn}(H), \tag{10}$$

$$Y = f_o(H\ W_{router})$$

where $H \in \mathbb{R}^{B \times N \times d}$, $B$ is the batch size, $N = 2$ is the number of experts, $d$ is the embedding dimension. $Router_{attn}$ is the self-attention mechanism[58] and $f_o$ is the output linear layer. This design enables MGT to adaptively fuse geometry-aware representations based on the relevance of each expert, thereby enhancing its performance across diverse downstream tasks.

**Pretraining**

We adopted a multi-objective SSL strategy that simultaneously performs denoising and contrastive learning across SE3 and SO3 graph representations. Specifically, for each crystal graph structure, we introduce random noise into geometric attributes, including angle and edge features in SE3 and SO3 graphs, respectively. The pretraining framework is trained to predict the perturbed noise directly from the corresponding graph embeddings, which can be expressed as:

$$\mathcal{L}_{SE3} = \sum_{i=1}^{m} \left[ \left\| \hat{\varepsilon}_i^\theta - (\tilde{\theta}_i - \theta_i) \right\|^2 \right] \tag{11}$$

$$\mathcal{L}_{SO3} = \sum_{i=1}^{m} \left[ \left\| \hat{\varepsilon}_i^e - (\tilde{e}_i - e_i) \right\|^2 \right] \tag{12}$$

where $\hat{\varepsilon}$ is the noise predicted by the model, $\theta$ and $e$ are the original attributes, $m$ is the batch size, $\tilde{\theta}$ and $\tilde{e}$ are the modified angle and edge features obtained by adding noise $\varepsilon$, which is sampled from a Gaussian distribution. In parallel, we empowered the model to capture shared geometric semantics through maximizing the mutual information between the SE3 and SO3 embeddings. This is achieved by normalizing the temperature-scaled cross-entropy (NT-Xent) loss[59]:

$$\mathcal{L}_{constrast} = -\frac{1}{2N} \sum_{k=1}^{N} \left[ \log \frac{e^{sim(z_i, z_j)/\tau}}{\sum_{j=1, j \neq i}^{2N} e^{sim(z_i, z_j)/\tau}} + \log \frac{e^{sim(z_j, z_i)/\tau}}{\sum_{j=1, j \neq i+N}^{2N} e^{sim(z_j, z_i)/\tau}} \right] \tag{13}$$

where $N$ is the batch size, $z_i$ and $z_j$ represent the graph embeddings from the SE3 and SO3, respectively. The cosine similarity between these two embeddings is $sim(z_i, z_j) = \frac{z_i\, z_j}{\|z_i\| \|z_j\|}$. $\tau$ is the temperature parameter, which scales the similarity values and controls the smoothness of the distribution. The pretraining objective integrates these components with balancing coefficients $\lambda_1$, $\lambda_2$ and $\lambda_3$:

$$\mathcal{L}_{total} = \lambda_1 \mathcal{L}_{constrast} + \lambda_2 \mathcal{L}_{SE3} + \lambda_3 \mathcal{L}_{SO3} \tag{14}$$

We employed the AdamW[60] optimizer with an initial learning rate of $10^{-5}$ for model pretraining with $\lambda_1 = 1.0$, $\lambda_2 = 0.5$ and $\lambda_3 = 0.5$. The learning rate was scheduled using cosine annealing with a predefined minimum value, gradually decreasing the learning rate over training epochs. Additionally, to enable that MGT can capture meaningful spatial structural information while avoiding excessive distribution shifts between clean and noisy conformations, the noise rate was set to 0.15. The warmup step was set to 10. The batch size was set to 128 and training epoch was set to 100. More detailed model pretraining settings are provided in Supplementary Table 7.

**Fine-tuning**

All downstream tasks, including transfer learning settings, were trained using the mean squared error (MSE) loss, defined as:

$$\mathcal{L}_{fine-tune} = \frac{1}{n}\sum_{i=1}^{n}(y_i - \hat{y}_i)^2 \qquad (15)$$

where $n$ is the batch size, $y_i$ and $\hat{y}_i$ denote the ground truth and predicted values of the $i$-th sample, respectively. We employed the AdamW optimizer with an initial learning rate of 0.0005 for downstream tasks and 0.006 for transfer learning. A cosine annealing scheduler was applied to gradually decay the learning rate throughout training. All models were trained for 500 epochs, with a batch size of 16 for downstream tasks and 8 for transfer learning tasks. More detailed fine-tuning settings are provided in Supplementary Table 8.

**Implementation**

The entire model pretraining and fine-tuning process was implemented using the open-source frameworks PyTorch[61] and PyTorch Geometric[62]. The data processing was conducted using the FairChem[63], ASE[64], pymatgen[65], and JARVIS-Tools[36] packages. All experiments were conducted on a 64-bit CentOS v8.0 server equipped with four NVIDIA RTX 4090 GPUs and 512 GB of RAM.

**Data availability**

The pretraining dataset OQMD used in this work is primarily sourced from the processed dataset provided by CrystaLLM (https://github.com/lantunes/CrystaLLM), following the commands and procedures outlined in their project documentation. Materials Project and JARVIS datasets are derived from the JARVIS-Tools package. For transfer learning, datasets such as Alloy-GMAE, FG-GMAE, OCD-GMAE, and OC20-LMAE are available at https://doi.org/10.5281/zenodo.12104162. HOIP dataset is available at https://datadryad.org/dataset/doi:10.5061/dryad.gq3rg. All the processed data resources are available at https://doi.org/10.5281/zenodo.15473642.

## Code availability

MGT is available at https://github.com/Turningl/MGT or https://doi.org/10.5281/zenodo.15473642.

## Acknowledgements

This work was supported by the China Ministry of Science and Technology (2020YFA0710203), the Joint Funds of the National Natural Science Foundation of China (U23A2081), the National Natural Science Foundation of China (22201271, 92261105 and 22221003), the Anhui Provincial Natural Science Foundation (2108085UD06 and 2208085UD04), the Anhui Provincial Key Research and Development Project (2023z04020010 and 2022a05020053), USTC Research Funds of the Double First-Class Initiative (YD2060002029 and YD2060006005), the Joint Funds from Hefei National Synchrotron Radiation Laboratory (KY2060000180 and KY2060000195) and the Fundamental Research Funds for the Central Universities (WK2060000088). The AI-driven experiments, simulations and model training were performed on the robotic AI-Scientist platform of Chinese Academy of Science.

## Authors contributions

Y.W. and K.C. conceived the study and supervised the research. L.Z. designed and implemented the computational framework, conducted benchmarks and case studies, and wrote the initial draft of the manuscript. Y.W., K.C., and L.Z. contributed to significant revisions. All authors discussed the results and provided feedback on the manuscript.

## Competing Interests

The authors declare no competing interests.

# References

1. Levi, Mikhael D., et al. Application of a quartz-crystal microbalance to measure ionic fluxes in microporous carbons for energy storage. *Nat. Mater.* **8**, 872-875 (2009).

2. Kim, Seok-Jin, et al. Structural control over single-crystalline oxides for heterogeneous catalysis. *Nat. Rev. Chem.* 1-18 (2025).

3. Xiang, L. et al. Synergistic Machine Learning Accelerated Discovery of Nanoporous Inorganic Crystals as Non-Absorbable Oral Drugs. *Adv. Mater.* **36**, 2404688 (2024).

4. Smoleński, T. et al. Signatures of Wigner crystal of electrons in a monolayer semiconductor. *Nature* **595**, 53–57 (2021).

5. Griesemer, S. D., Xia, Y. & Wolverton, C. Accelerating the prediction of stable materials with machine learning. *Nat Comput Sci* **3**, 934–945 (2023).

6. Gusev, V. V. et al. Optimality guarantees for crystal structure prediction. *Nature* **619**, 68–72 (2023).

7. Deng, B. et al. CHGNet as a pretrained universal neural network potential for charge-informed atomistic modelling. *Nat Mach Intell* **5**, 1031–1041 (2023).

8. Reed, J. & Ceder, G. Role of Electronic Structure in the Susceptibility of Metastable Transition-Metal Oxide Structures to Transformation. *Chem. Rev.* **104**, 4513–4534 (2004).

9. Sibanda, D., Oyinbo, S. T. & Jen, T.-C. A review of atomic layer deposition modelling and simulation methodologies: Density functional theory and molecular dynamics. *NANOTECHNOL. REV.* **11**, 1332–1363 (2022).

10. Xie, T. & Grossman, J. C. Crystal Graph Convolutional Neural Networks for an Accurate and Interpretable Prediction of Material Properties. *Phys. Rev. Lett.* **120**, 145301 (2018).

11. Yan, K., Liu, Y., Lin, Y. & Ji, S. Periodic Graph Transformers for Crystal Material Property Prediction. *Adv. Neural. Inf. Process. Syst.* **35**, 15066–15080 (2022).

12. Lin, Y. et al. Efficient Approximations of Complete Interatomic Potentials for Crystal Property Prediction. In *International Conference on Machine Learning*, 21260-21287 (2023).

13. Huang, J., Xing, Q., Ji, J. & Yang, B. PerCNet: Periodic complete representation for crystal graphs. *Neural Networks* **181**, 106841 (2025).

14. Yan, K, et al. Complete and efficient graph transformers for crystal material property prediction. In *International Conference on Learning Representations*, (2024).

15. Batzner, S. et al. E(3)-equivariant graph neural networks for data-efficient and accurate interatomic potentials. *Nat Commun* **13**, 2453 (2022).

16. Liao, Y.-L. & Smidt, T. Equiformer: Equivariant Graph Attention Transformer for 3D Atomistic Graphs. In *International Conference on Learning Representations*, (2022).

17. Liao, Y.-L., et al. Equiformerv2: Improved equivariant transformer for scaling to higher-degree representations. In *International Conference on Learning Representations*, (2024).

18. Jiang, X., Tan, L. & Zou, Q. DGCL: dual-graph neural networks contrastive learning for molecular property prediction. *Briefings in Bioinformatics* **25**, bbae474 (2024).

19. Rizve, M. N., et al. Exploring complementary strengths of invariant and equivariant representations for few-shot learning. In *Proceedings of the IEEE/CVF Conference on Computer Vision and Pattern Recognition*, 10836–10846 (CVPR, 2021).

20. Liu, Y. et al. RoBERTa: A Robustly Optimized BERT Pretraining Approach. Preprint at https://arxiv.org/abs/1907.11692 (2019).

21. Ding, N. et al. Parameter-efficient fine-tuning of large-scale pre-trained language models. *Nat Mach Intell* **5**, 220–235 (2023).

22. Doersch, C. & Zisserman, A. Multi-task Self-Supervised Visual Learning. In *Proceedings of the IEEE international conference on computer vision*, 2051-2060 (ICCV, 2017).

23. Saeed, A., Ozcelebi, T. & Lukkien, J. Multi-task Self-Supervised Learning for Human Activity Detection. *Proc. ACM Interact. Mob. Wearable Ubiquitous Technol.* **3**, 1–30 (2019).

24. Jordan, M. I., & Jacobs, R. A. Hierarchical mixtures of experts and the EM algorithm. *Neural computation* **6**, 181-214 (1994).

25. Liu, A., et al. Deepseek-v3 technical report. Preprint at https://arxiv.org/abs/2412.19437 (2024).

26. Chen, J., Huang, X., Hua, C., He, Y. & Schwaller, P. A multi-modal transformer for predicting global minimum adsorption energy. *Nat Commun* **16**, 3232 (2025).

27. Kim, C., Huan, T. D., Krishnan, S. & Ramprasad, R. A hybrid organic-inorganic perovskite dataset. *Sci Data* **4**, 170057 (2017).

28. Van der Maaten, L. & Hinton, G. Visualizing data using t-SNE. *J. Mach. Learn. Res.* **9**, 2579–2605 (2008).

29. Xie, T., et al. Crystal diffusion variational autoencoder for periodic material generation. Preprint

at https://arxiv.org/abs/2110.06197 (2021).

30. Fang, X. *et al.* Geometry-enhanced molecular representation learning for property prediction. *Nat Mach Intell* **4**, 127–134 (2022).

31. Rong, Y. et al. Self-supervised graph transformer on large-scale molecular data. *Adv. Neural Inf. Process. Syst.* **33**, 12559–12571 (2020).

32. Hu, W. et al. Strategies for pre-training graph neural networks. In *8th International Conference on Learning Representations* (ICLR, 2020).

33. Vincent, P., Larochelle, H., Bengio, Y. & Manzagol, P.-A. Extracting and composing robust features with denoising autoencoders. In *ICML '08: Proc. of the 25th International Conference on Machine Learning* 1096–1103 (ACM, 2008).

34. Wang, Y., Wang, J., Cao, Z. & Barati Farimani, A. Molecular contrastive learning of representations via graph neural networks. *Nat Mach Intell* **4**, 279–287 (2022).

35. Chen, C., Ye, W., Zuo, Y., Zheng, C. & Ong, S. P. Graph Networks as a Universal Machine Learning Framework for Molecules and Crystals. *Chem. Mater.* **31**, 3564–3572 (2019).

36. Choudhary, K. *et al.* The joint automated repository for various integrated simulations (JARVIS) for data-driven materials design. *npj Comput Mater* **6**, 173 (2020).

37. Choudhary, K. & DeCost, B. Atomistic Line Graph Neural Network for improved materials property predictions. *npj Comput Mater* **7**, 185 (2021).

38. Nie, J., Xiao, P., Ji, K. & Gao, P. ReGNet: Reciprocal Space-Aware Long-Range Modeling and Multi-Property Prediction for Crystals. Preprint at https://arxiv.org/abs/2502.02748 (2025).

39. Montavon, G., Samek, W. & Müller, K.-R. Methods for interpreting and understanding deep neural networks. *Digital Signal Processing* **73**, 1–15 (2018).

40. Weiss, K., Khoshgoftaar, T. M. & Wang, D. A survey of transfer learning. *J Big Data* **3**, 9 (2016).

41. Dai, W., Yang, Q., Xue, G.-R. & Yu, Y. Boosting for transfer learning. *In ICML '07: Proc. of the 24th International Conference on Machine Learning 193–200.* (ACM, 2008).

42. Butler, K. T., Frost, J. M., Skelton, J. M., Svane, K. L. & Walsh, A. Computational materials design of crystalline solids. *Chem. Soc. Rev.* **45**, 6138–6146 (2016).

43. Chen, C. *et al.* Accelerating Computational Materials Discovery with Machine Learning and Cloud High-Performance Computing: from Large-Scale Screening to Experimental Validation. *J.*

*Am. Chem. Soc.* **146**, 20009–20018 (2024).

44. Lin, D.-Z. *et al.* A high-throughput experimentation platform for data-driven discovery in electrochemistry. *Sci. Adv.* **11**, eadu4391 (2025).

45. Szymanski, N. J. *et al.* An autonomous laboratory for the accelerated synthesis of novel materials. *Nature* **624**, 86–91 (2023).

46. Gainza, P. *et al.* De novo design of protein interactions with learned surface fingerprints. *Nature* **617**, 176–184 (2023).

47. Wang, Z. & You, F. Leveraging generative models with periodicity-aware, invertible and invariant representations for crystalline materials design. *Nat Comput Sci*, 1-12, (2025).

48. Jiao, R. et al. Crystal structure prediction by joint equivariant diffusion. *Adv. Neural Inf. Process. Syst.* **36**, 17464-17497 (2023).

49. Saal, J. E., Kirklin, S., Aykol, M., Meredig, B. & Wolverton, C. Materials Design and Discovery with High-Throughput Density Functional Theory: The Open Quantum Materials Database (OQMD). *JOM* **65**, 1501–1509 (2013).

50. Antunes, L. M., Butler, K. T. & Grau-Crespo, R. Crystal structure generation with autoregressive large language modeling. *Nat Commun* **15**, 10570 (2024).

51. Mamun, O., Winther, K. T., Boes, J. R. & Bligaard, T. High-throughput calculations of catalytic properties of bimetallic alloy surfaces. *Sci Data* **6**, 76 (2019).

52. Pablo-García, S. *et al.* Fast evaluation of the adsorption energy of organic molecules on metals via graph neural networks. *Nat Comput Sci* **3**, 433–442 (2023).

53. Lan, J. *et al.* AdsorbML: a leap in efficiency for adsorption energy calculations using generalizable machine learning potentials. *npj Comput Mater* **9**, 172 (2023).

54. Chanussot, L. *et al.* Open Catalyst 2020 (OC20) Dataset and Community Challenges. *ACS Catal.* **11**, 6059–6072 (2021).

55. Acosta, F. M. A. Radial basis function and related models: An overview. *Signal Processing* **45**, 37–58 (1995).

56. Thomas, N. *et al.* Tensor field networks: Rotation- and translation-equivariant neural networks for 3D point clouds. Preprint at https://arxiv.org/abs/1802.08219 (2018).

57. Geiger, M. & Smidt, T. e3nn: Euclidean Neural Networks. Preprint at

https://arxiv.org/abs/2207.09453 (2022).

58. Vaswani, A. et al. Attention is all you need. *Adv. Neural Inf. Process. Syst.* **30**, 5998–6008 (2017).

59. Sohn, K. Improved deep metric learning with multi-class n-pair loss objective. *Adv. Neural Inf. Process. Syst.* **29** (2016).

60. Zhuang, Z., Liu, M., Cutkosky, A. & Orabona, F. Understanding AdamW through Proximal Methods and Scale-Freeness. Preprint at https://arxiv.org/abs/2202.00089 (2022).

61. Paszke, A. et al. PyTorch: An Imperative Style, High-Performance Deep Learning Library. *Adv. Neural Inf. Process. Syst.* **32**, 8024–8035 (2019).

62. Fey, M. & Lenssen, J. E. Fast Graph Representation Learning with PyTorch Geometric. Preprint at http://arxiv.org/abs/1903.02428 (2019).

63. Barroso-Luque, L. *et al.* Open Materials 2024 (OMat24) Inorganic Materials Dataset and Models. Preprint at https://arxiv.org/abs/2410.12771 (2024).

64. Hjorth Larsen, A. *et al.* The atomic simulation environment—a Python library for working with atoms. *J. Phys.: Condens. Matter* **29**, 273002 (2017).

65. Ong, S. P. *et al.* Python Materials Genomics (pymatgen): A robust, open-source python library for materials analysis. *Computational Materials Science* **68**, 314–319 (2013).

66. Cui, T. *et al.* Geometry-enhanced pretraining on interatomic potentials. *Nat Mach Intell* **6**, 428–436 (2024).

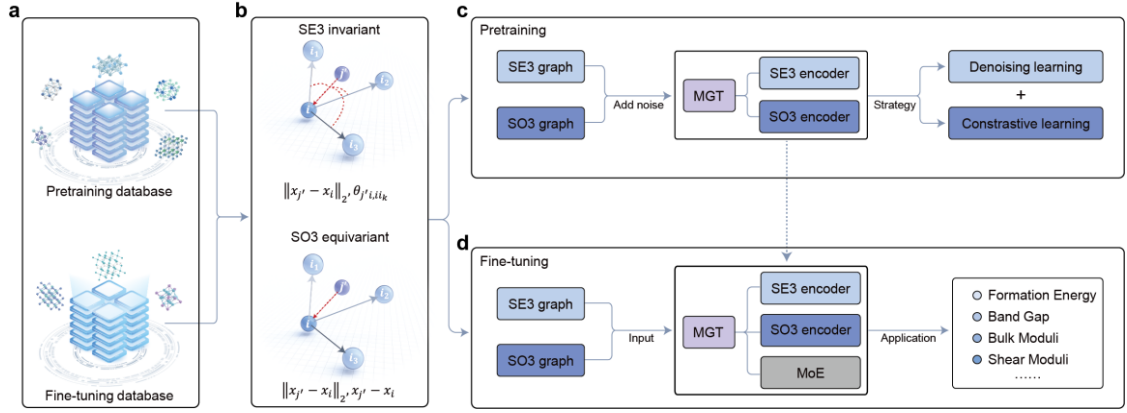

**Fig. 1. Schematic diagram of MGT framework workflow. a**, database preparation, which are pretraining database and fine-tuning database, respectively. **b**, SE3 invariant features include the edge length $\|x_{j\prime} - x_i\|_2$ and the angle $\theta_{j\prime i, i i_k}$, $k \in \{1, 2, 3\}$. SO3 equivariant features include the edge length $\|x_{j\prime} - x_i\|_2$ and the equivariant vectors $x_{j\prime} - x_i$. **c**, For pretraining, using multi-objective SSL strategy that simultaneously performs denoising and contrastive learning. **d**, For fine-tuning, MoE is employed to adaptively adjust the weight assigned to SE3 and SO3 embeddings based on the specific target task.

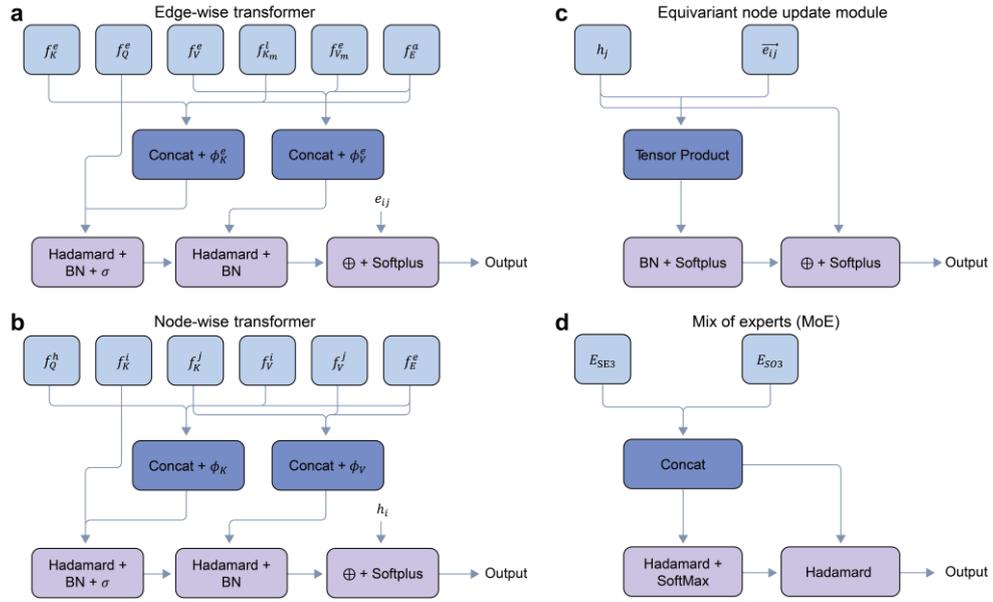

**Fig. 2. MGT architecture, including multiple different modules. a**, Edge-wise transformer. **b**, Node-wise transformer. **c**, Equivariant node update module. **d**, Mix of experts (MoE), using self-attention mechanism.

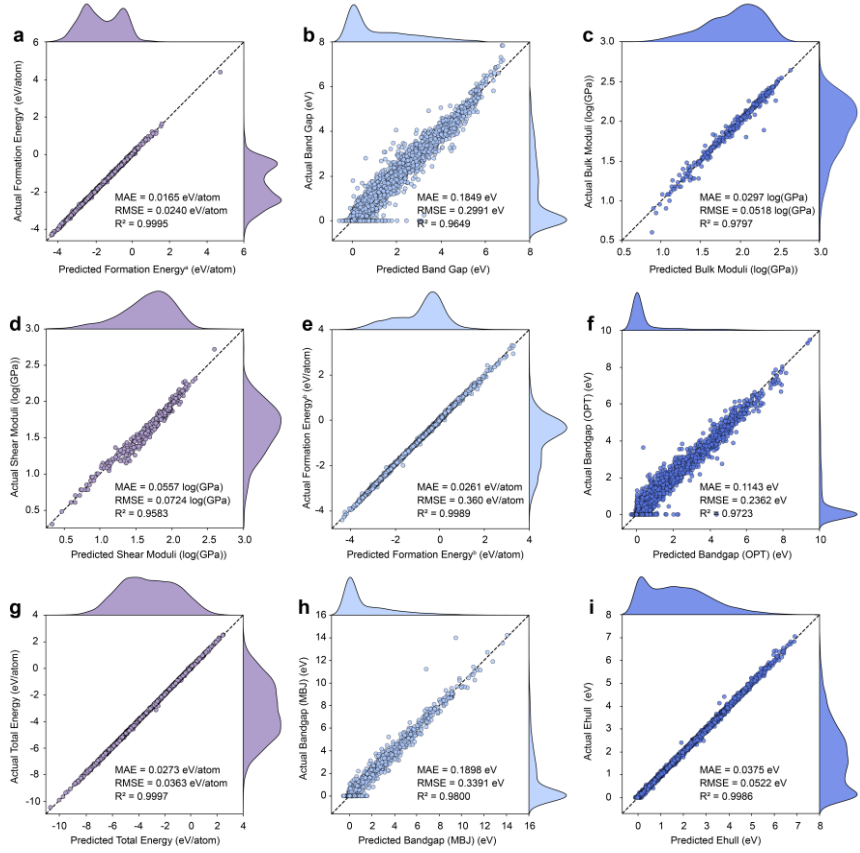

**Fig. 3. Performance comparison of MGT on Materials Project and JARVIS datasets with MAE, RMSE and R-squared value ($R^2$). a-d,** Actual and predicted results of Formation Energy, Band Gap, Bulk Moduli and Shear Moduli in Materials Project dataset. **e-i**, Actual and predicted results of Formation Energy, Bandgap (OPT), Total Energy, Bandgap (MBJ) and Ehull in JARVIS dataset.

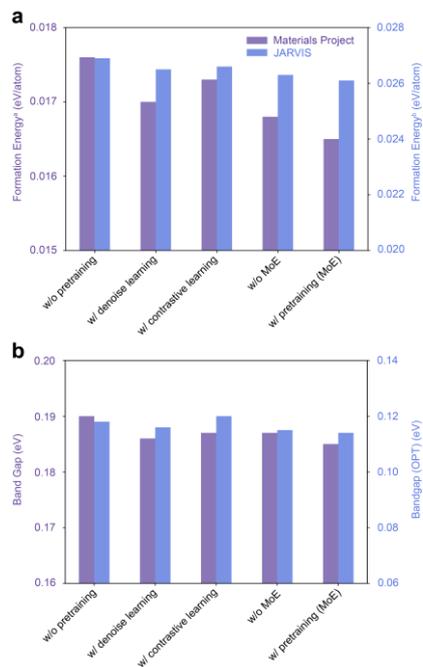

**Fig. 4. Ablation experiment results of MGT with different training strategies. w/o: without. w/: with. a,** Performance comparison of MGT in the Formation Energy prediction task for Materials Project and JARVIS datasets, respectively. **b,** Performance comparison of MGT in the Band Gap and Bandgap (OPT) prediction tasks for Materials Project and JARVIS datasets, respectively.

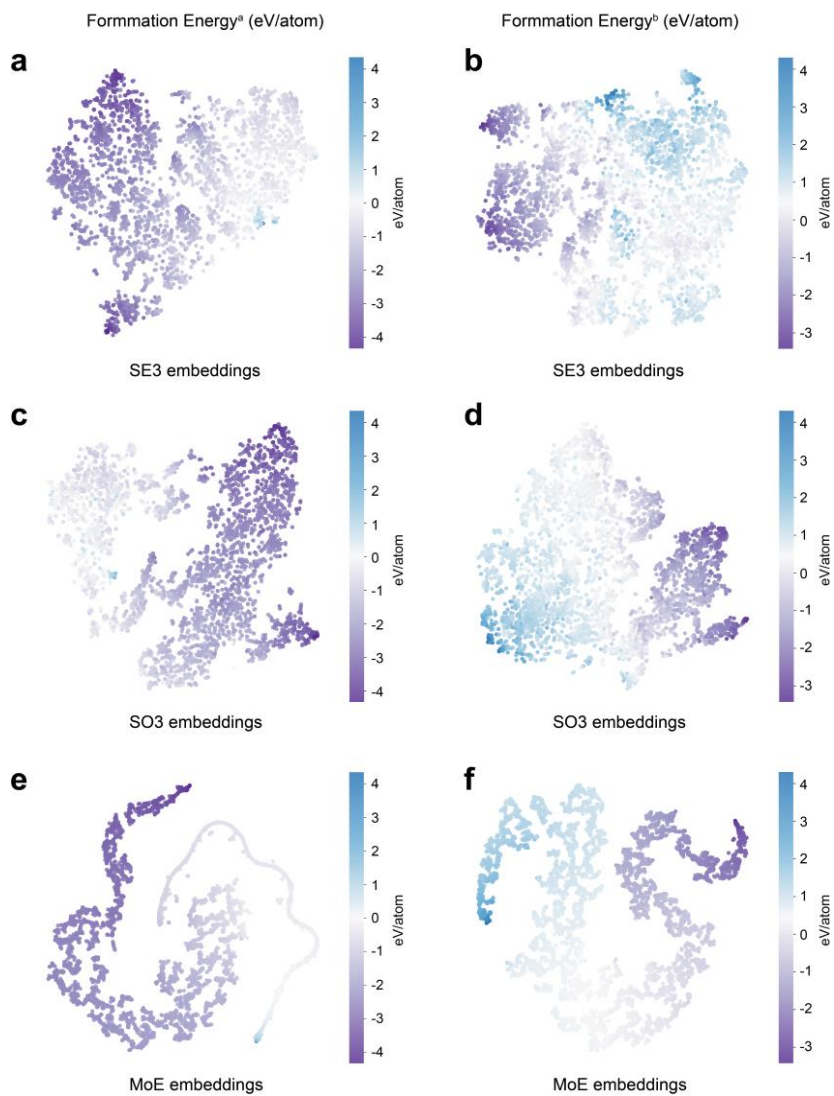

**Fig. 5. Visualizations of the Formation Energy prediction task for Materials Project and JARVIS datasets. a,** SE3 embeddings for Materials Project dataset. **b,** SE3 embeddings for JARVIS dataset. **c,** SO3 embeddings for Materials Project dataset. **d,** SO3 embeddings for JARVIS dataset. **e,** MoE embeddings for Materials Project dataset. **f,** MoE embeddings for JARVIS dataset.

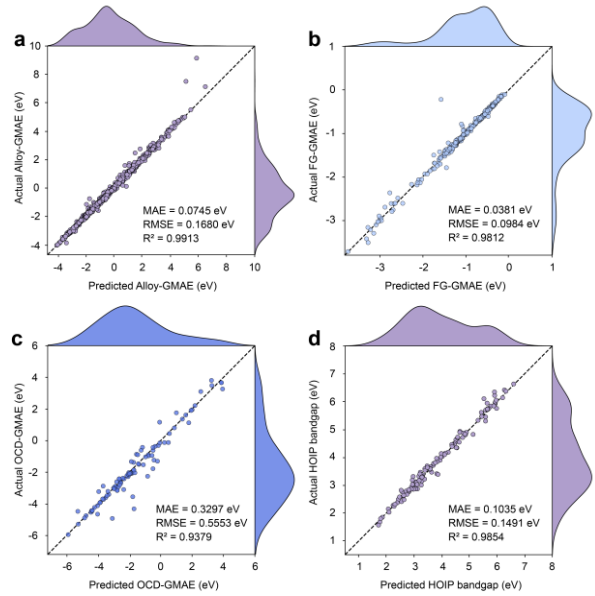

**Fig. 6. Performance comparison of MGT on Alloy-GMAE, FG-GMAE, OCD-GMAE and HOIP datasets with MAE, RMSE, and $R^2$. a**, Actual and predicted results of Alloy-GMAE prediction task. **b**, Actual and predicted results of FG-GMAE prediction task. **c**, Actual and predicted results of OCD-GMAE prediction task. **d**, Actual and predicted results of HOIP bandgap prediction task.

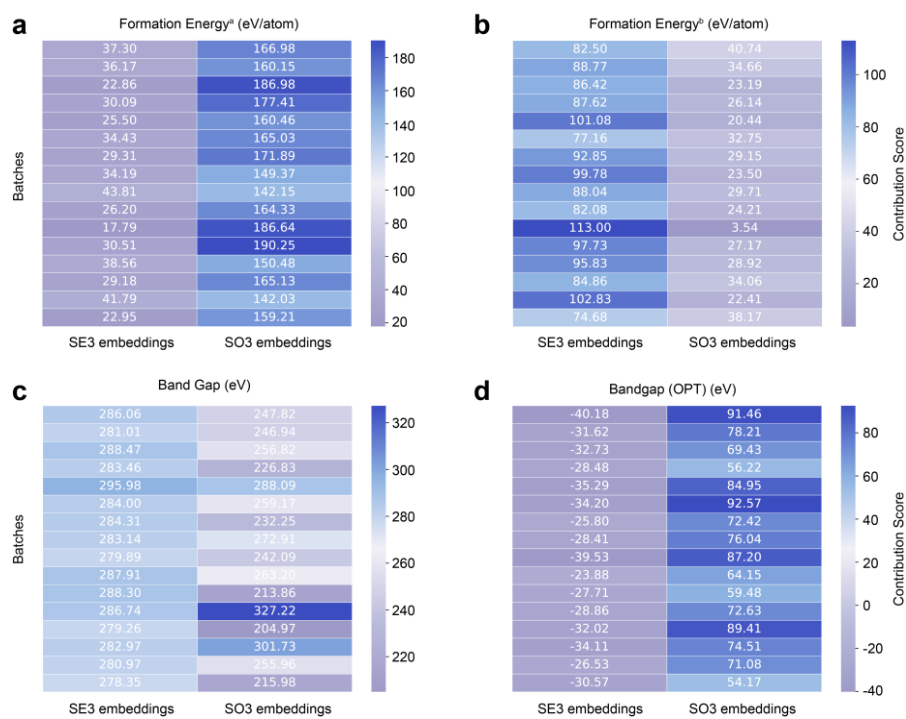

**Extended Data Fig. 1. Contribution scores of SE3 and SO3 embeddings in the MoE module for Materials Project and JARVIS datasets with batches of size 16. a,** the Formation Energy prediction task for Materials Project dataset. **b,** the Formation Energy prediction task for JARVIS dataset. **c,** the Band Gap prediction task for Materials Project dataset. **d,** the Bandgap (OPT) prediction task for Materials Project dataset.

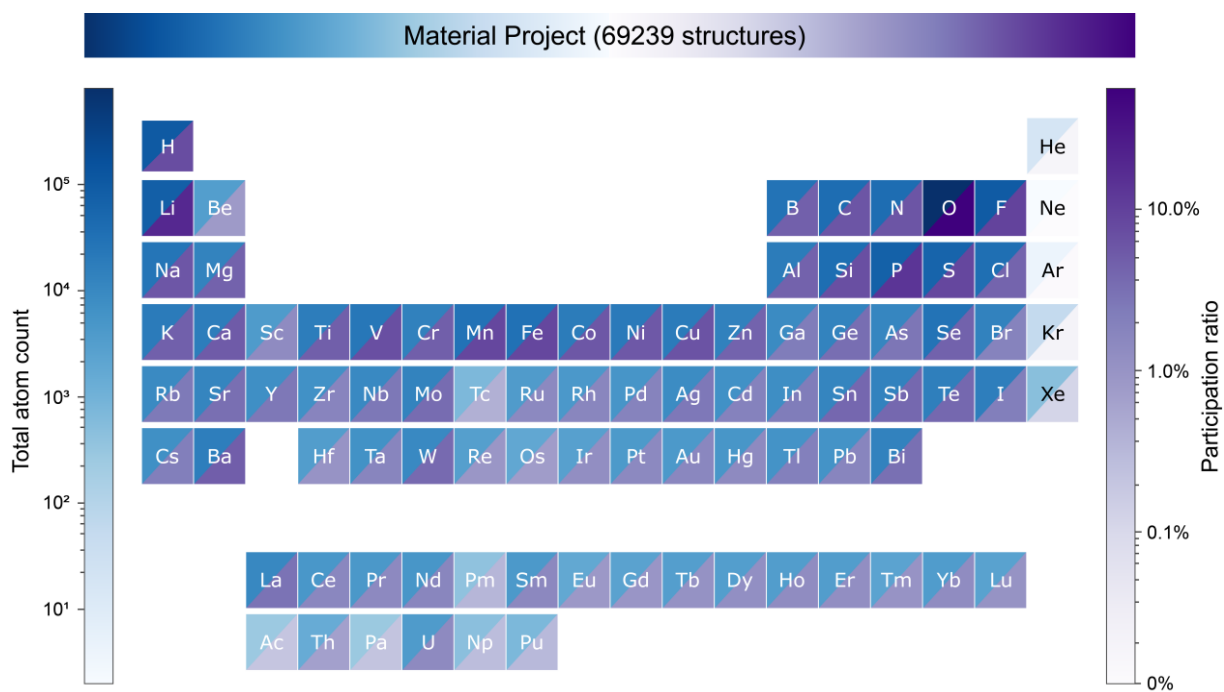

**Extended Data Fig. 2. Element distribution of the Material Project dataset.** The colour on the upper-left triangle indicates the total number of atoms of an element. The colour on the lower-right indicates the proportion of structures in which the element appears, representing its structural participation ratio.

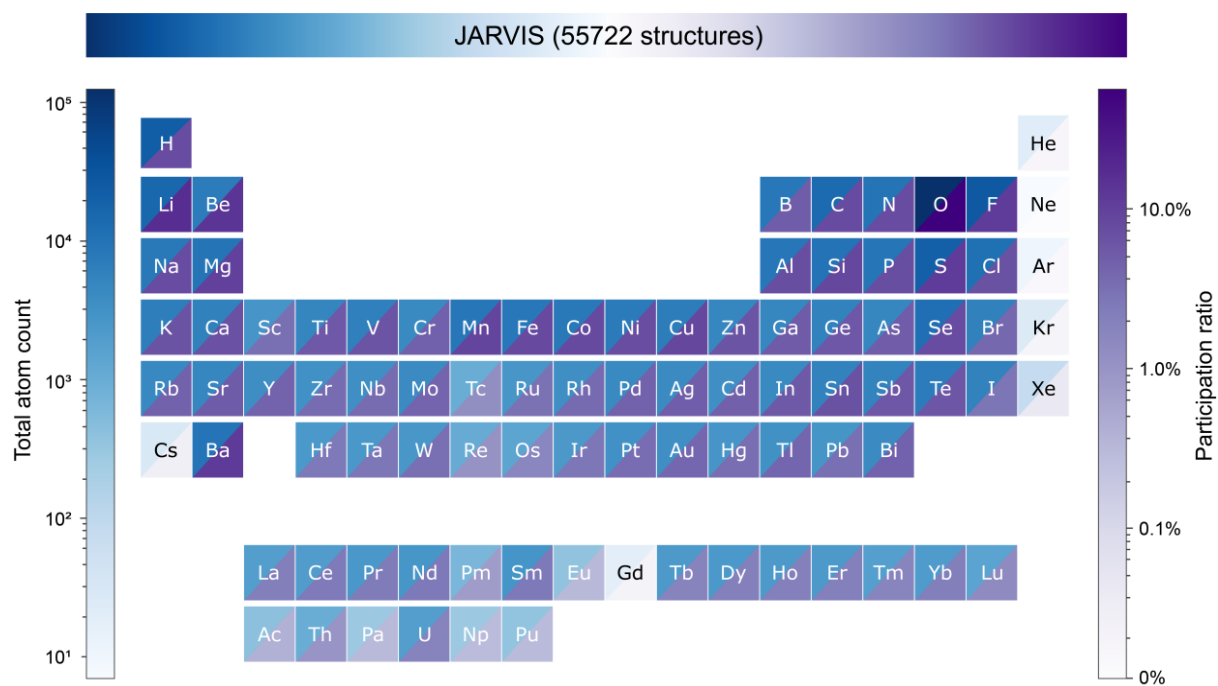

**Extended Data Fig. 3. Element distribution of the JARVIS dataset.** The colour on the upper-left triangle indicates the total number of atoms of an element. The colour on the lower-right indicates the proportion of structures in which the element appears, representing its structural participation ratio.

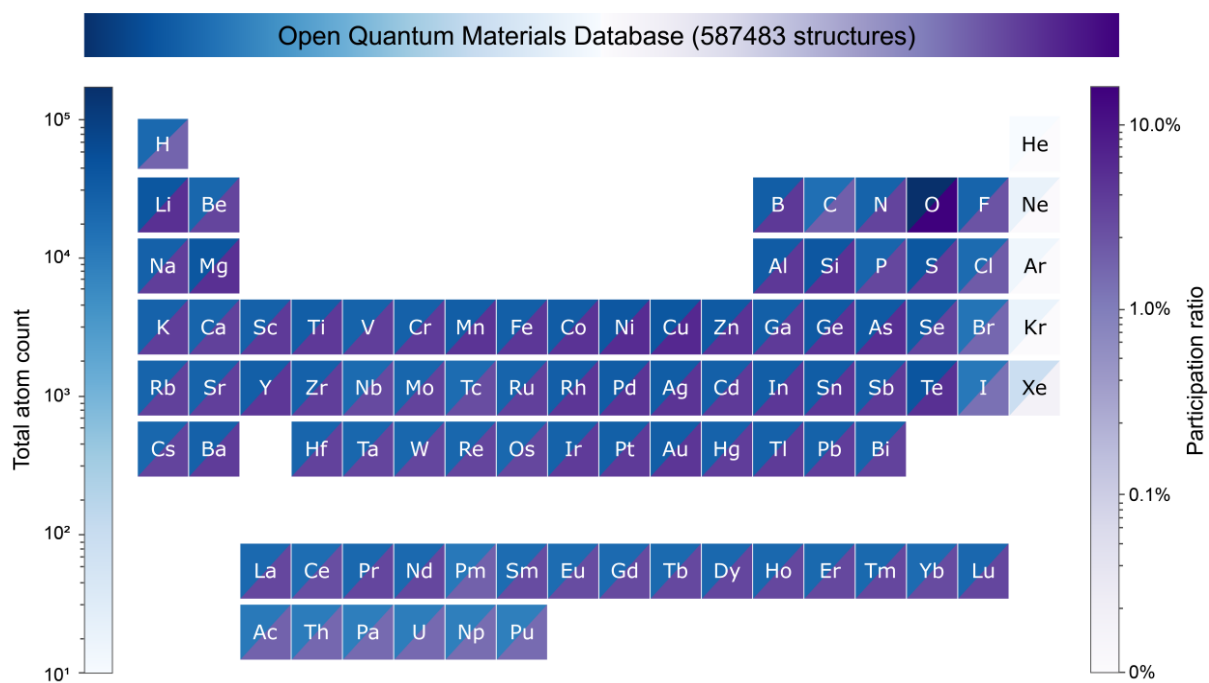

**Extended Data Fig. 4. Element distribution of the OQMD Dataset.** The colour on the upper-left triangle indicates the total number of atoms of an element. The colour on the lower-right indicates the proportion of structures in which the element appears, representing its structural participation rat